\documentclass[11pt]{article}

\usepackage[preprint]{acl}

\usepackage{times}
\usepackage{latexsym}

\usepackage[T1]{fontenc}

\usepackage[utf8]{inputenc}

\usepackage{microtype}

\usepackage{inconsolata}

\usepackage{graphicx}
\usepackage{booktabs}
\usepackage{subcaption}
\usepackage{placeins}

\title{How Do LLMs Understand Legal Philosophy? \\
Diagnosing Legal-Philosophical Reasoning in LLMs}

\author{
  Shuyuan Zheng\textsuperscript{1} \quad
  Yun-seo Lee\textsuperscript{3} \quad
  Xing Gao\textsuperscript{4} \quad
  Jialu Wu\textsuperscript{4} \quad
  Jiabao Hu\textsuperscript{5} \\
  \textbf{Katrina Qiyao Wang}\textsuperscript{6} \quad
  \textbf{Xinbo Lin}\textsuperscript{7} \quad
  \textbf{Qianying Liu}\textsuperscript{8} \quad
  \textbf{Chuan Xiao}\textsuperscript{1} \quad
  \textbf{Tao Huang}\textsuperscript{2}\thanks{Corresponding author.} \\
  {\normalfont\small
  \textsuperscript{1}The University of Osaka \hspace{0.5em}
  \textsuperscript{2}City University of Hong Kong \hspace{0.5em}
  \textsuperscript{3}UC Davis \hspace{0.5em}
  \textsuperscript{4}UC Berkeley} \\
  {\normalfont\small
  \textsuperscript{5}Peking University \hspace{0.5em}
  \textsuperscript{6}University of Wisconsin--Madison \hspace{0.5em}
  \textsuperscript{7}Shanghai Jiao Tong University \hspace{0.5em}
  \textsuperscript{8}NII LLMC} \\
  {\normalfont\small
  \texttt{zheng@ist.osaka-u.ac.jp} \quad
  \texttt{taohuang@cityu.edu.hk}}
}

\input{savespace.tex}

\usepackage{enumitem}
\setlist[itemize]{leftmargin=0pt, itemindent=0pt, labelsep=0.5em}
\setlist[enumerate]{leftmargin=0pt, itemindent=0pt, labelsep=0.5em}

\SetEnumitemKey{questionnaireoptions}{labelindent=0.5em, labelwidth=0.5em,
  leftmargin=!, itemindent=0pt, labelsep=0.5em, align=left}


\usepackage{soul}
\usepackage{tcolorbox}
\tcbuselibrary{skins,breakable}

\definecolor{lightred}{RGB}{255, 200, 200}
\definecolor{lightgreen}{RGB}{200, 255, 200}
\definecolor{deepviolet}{RGB}{96, 48, 96}

\tcbset{
  promptbox/.style={
    breakable,
    top=10pt,
    colback=white,
    colframe=violet,
    colbacktitle=deepviolet,
    fontupper=\itshape,
    enhanced,
    center,
    attach boxed title to top left={yshift=-0.1in,xshift=0.15in},
    boxed title style={boxrule=0pt,colframe=white},
  }
}
\newtcolorbox{promptframed}[2][]{promptbox,title=#2,#1}

\tcbset{
  promptbox1/.style={
    colback=white,
    colframe=violet,
    fontupper=\itshape,
    enhanced,
    center,
  }
}
\newtcolorbox{promptframed1}[2][]{promptbox1,title=#2,#1}

\newcounter{promptcounter}
\newenvironment{prompt}[1]
{
  \refstepcounter{promptcounter}\par\medskip
  \begin{promptframed}{Prompt~\thepromptcounter: #1}
}
{
  \end{promptframed}
}

\begin{document}
\maketitle

\begin{abstract}
Legal reasoning is a fundamental capability for large language models (LLMs) to complete legal AI tasks such as legal judgment prediction. In legal practice, this reasoning is shaped by legal philosophy: a judge's view of the nature and sources of law guides how legal materials are interpreted and applied, so judges working from the same statutes and precedents can reach different judgments. Existing research has largely overlooked whether LLMs can understand legal philosophy and apply it in legal reasoning, an ability we call \textit{legal-philosophical reasoning}. To fill this gap, we introduce LePhilBench, a benchmark for evaluating the legal-philosophical reasoning of LLMs that comprises two benchmarking tasks and a dataset. The two tasks evaluate how well LLMs understand legal philosophy and how well they apply it in legal reasoning, respectively. The dataset is manually and meticulously constructed. Its test set contains 220 U.S. Supreme Court opinions. A law professor designed fourteen annotation questions on legal philosophy, and three law-trained annotators with formal training in jurisprudence answered all fourteen questions for every opinion, yielding 9,240 legal-philosophical labels. LePhilBench takes the first step toward exploring legal-philosophical reasoning in LLMs.\footnote{We will publicly release LePhilBench upon acceptance.}
\end{abstract}

\begin{figure*}[t]
    \centering
    \includegraphics[width=0.9\textwidth]{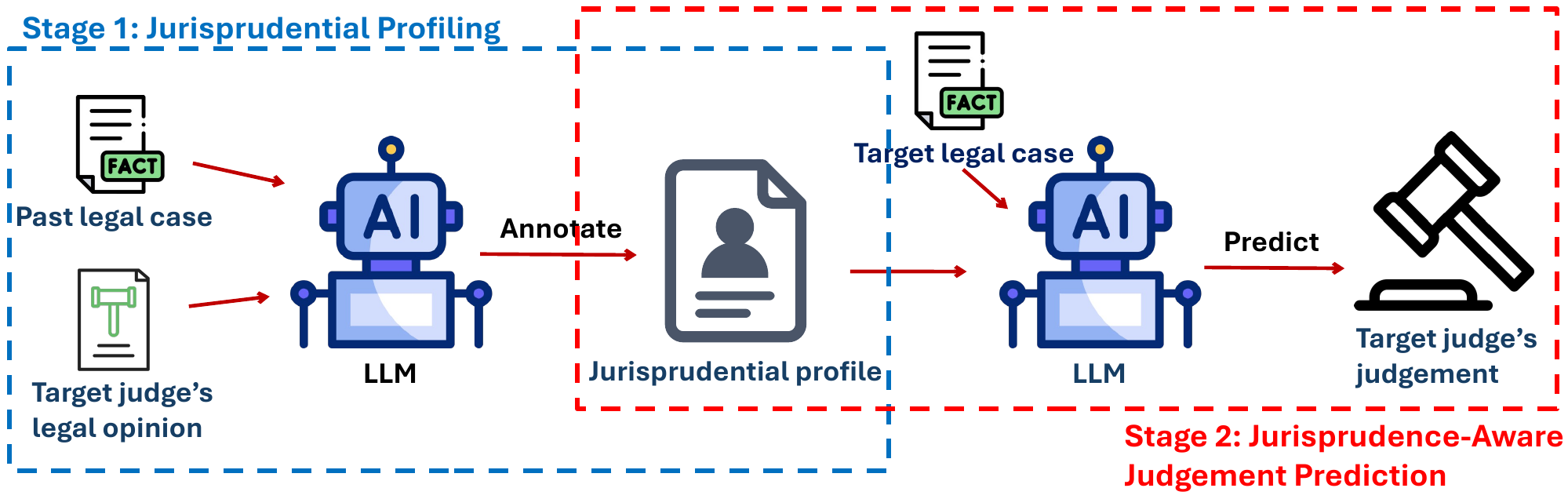}
    \caption{Overview of jurisprudential profiling (JP) and jurisprudence-aware judgment prediction (JAJP): inferring a jurisprudential profile from a judge's past opinion(s) and using it to predict the judge's judgment in a target case.}
    \label{fig:task-overview}
\end{figure*}

\section{Introduction}
\label{sec:introduction}

Legal reasoning, the ability to apply legal rules and principles to the facts of a particular case to reach a legal decision, is a central capability target for large language models (LLMs) in the legal domain. It underlies tasks such as legal judgment prediction \citep{fei-etal-2024-lawbench}, legal fact prediction \citep{liu-etal-2025-legal}, and legal opinion generation \citep{huang-etal-2023-lawyer}. In each of these tasks, a model must apply abstract legal norms to concrete case facts to reach a legal decision. Whether LLMs can perform this reasoning reliably therefore determines how far their legal decisions can be trusted.

In real-world legal practice, however, legal reasoning is not a mechanical application of rules to facts. It is shaped by legal philosophy. How legal professionals understand the nature of law and the sources of its legitimacy influences how they interpret legal materials and apply them to a case. As a result, judges who work from the same statutes and precedents can reach different judgments because they hold different jurisprudential views, for example on the relative weight of text, purpose, precedent, and moral principle. Existing evaluations and benchmarks of LLMs' legal reasoning have largely overlooked this influence. They mainly assess task correctness and law-to-fact reasoning \citep{shui-etal-2023-comprehensive,jiang-yang-2023-legal}. Some benchmarks cover related abilities, such as interpretive-tool recognition in LegalBench \citep{guha-etal-2023-legalbench} and value-based prosecutorial discretion in PDP-Bench \citep{lu-etal-2026-cases}. Even these do not examine whether a model's legal reasoning reflects a jurisprudential position. An unresolved research question therefore remains: \textit{can LLMs engage in legal-philosophical reasoning as legal professionals do, i.e., understand legal philosophy and use it to personalize legal reasoning?}

To fill this gap, this paper explores how LLMs can perform legal-philosophical reasoning. We construct LePhilBench, a benchmark framework consisting of two tasks (see Figure~\ref{fig:task-overview}): \textit{jurisprudential profiling} (JP) and \textit{jurisprudence-aware judgment prediction} (JAJP). JP asks an LLM to read a human judge's legal opinions and annotate the jurisprudential tendencies they may stand along fourteen dimensions, thereby constructing the judge's jurisprudential profile. JAJP asks an LLM to predict, based on a target judge's jurisprudential profile, how that judge would resolve the disputed legal question in a given case. This framework is inspired by a common practice in litigation strategy. Before arguing a case, lawyers often read the past opinions of the judge to form an impression of their legal-philosophical tendencies, and then draw on that impression to anticipate how the judge is likely to decide the key legal questions in the case. LePhilBench makes this otherwise tacit impression explicit as a structured jurisprudential profile: JP mirrors the step of forming it, and JAJP mirrors the step of drawing on it.

To drive these two tasks, we construct the LePhilBench dataset. We collect the opinions of the Court and the dissenting opinions from 489 U.S. Supreme Court cases decided over the past two decades, and use them to construct the jurisprudential profiles of eleven target justices. To make these profiles explicit, a law professor specializing in U.S. constitutional law helped us design fourteen dimensions of jurisprudential tendency. Each dimension measures where a judge stands between a pair of competing jurisprudential views, such as Natural Law vs. Legal Positivism. We then select 110 cases as the test set. Three law-trained annotators with formal training in jurisprudence annotate the opinion of the Court and the dissenting opinion in each test case, 220 opinions in total. For each opinion, each annotator provides individual labels for the fourteen tendencies and for the opinion's judgement of the case's key legal question, yielding $110 \times 2 \times 14 \times 3 = 9{,}240$ jurisprudential labels.

Our evaluation on LePhilBench shows that frontier LLMs have considerable potential for legal-philosophical reasoning. First, in JP, the agreement between frontier LLMs and law-trained human annotators is close to the agreement among the human annotators themselves. Frontier LLMs therefore recover the jurisprudential tendencies expressed in judicial opinions about as consistently as law-trained human annotators do. This suggests that they can \textit{understand} legal philosophy as law-trained humans do. Second, in JAJP, frontier LLMs predict how a target judge would resolve a case's key legal question with fairly high accuracy, even though the judge's name is withheld and only their jurisprudential profile is provided. Because the judge's identity is hidden, a correct prediction can be attributed to the reasoning about the jurisprudential tendencies recorded in the profile. This suggests that frontier LLMs have the strong potential to \textit{apply} legal philosophy as legal professionals do.

In summary, this paper makes three contributions. \textbf{(1)} We propose LePhilBench, a benchmark framework that evaluates the legal-philosophical reasoning of LLMs through two complementary tasks, JP and JAJP. \textbf{(2)} We construct the LePhilBench dataset from 489 U.S. Supreme Court cases, with 9{,}240 annotations carefully produced by law-trained annotators, whose case answers are verified by a law professor. \textbf{(3)} We conduct extensive evaluations of frontier LLMs on LePhilBench, revealing both their capabilities and their limitations in legal-philosophical reasoning. We hope LePhilBench can serve as a starting point for building legal AI systems that reason not only with legal rules but also with legal philosophy.

\section{The LePhilBench Dataset}
\label{sec:lephilbench}

\begin{table}[!t]
\centering
\small
\renewcommand{\arraystretch}{1.1}
\setlength{\tabcolsep}{4pt}
\begin{tabular}{@{}p{0.30\columnwidth}p{0.66\columnwidth}@{}}
\hline
\textbf{Information} & \textbf{Description}\tabularnewline
\hline
\multicolumn{2}{@{}l}{\textbf{Case record}}\tabularnewline
Date (Oyez) & \raggedright Date the Court decided the case\tabularnewline
Case facts (Oyez) & \raggedright Summary of the facts of the case\tabularnewline
\textcolor{blue}{Key legal question (Oyez \& us)} & \raggedright Binary question at the center of the dispute\tabularnewline
\hline
\multicolumn{2}{@{}l}{\textbf{Opinion record}}\tabularnewline
Type (Justia) & \raggedright Court Opinion or Dissent\tabularnewline
Author (Justia) & \raggedright Justice who wrote the opinion\tabularnewline
Main text (Justia) & \raggedright Full text of the opinion\tabularnewline
\textcolor{blue}{Exclusive joiners (us)} & \raggedright Justices who fully joined the opinion and neither wrote nor joined any other merits opinion in the case\tabularnewline
\textcolor{blue}{Case answer (us)} & \raggedright Gold answer to the key legal question (A = Yes, B = No)\tabularnewline
\textcolor{blue}{Tendency labels for Q1--Q14 (us)} & \raggedright Three individual annotators' labels for each jurisprudential tendency\tabularnewline
\hline
\end{tabular}
\caption{Key fields in LePhilBench. The source of each field is given in parentheses.}
\label{tab:dataset-records}
\end{table}

\begin{table}[t]
\centering
\small
\begin{tabular}{@{}p{0.46\columnwidth}rrr@{}}
\hline
Statistic & \multicolumn{1}{c}{Court} & \multicolumn{1}{c}{Dissent} & \multicolumn{1}{c}{Overall} \\
\hline
\multicolumn{4}{@{}l}{\textit{Full set}} \\
Cases & -- & -- & 489 \\
Opinions & 489 & 489 & 978 \\
Authors & 15 & 15 & 15 \\
\hline
\multicolumn{4}{@{}l}{\textit{Test set}} \\
Cases & -- & -- & 110 \\
Opinions & 110 & 110 & 220 \\
Authors & 14 & 14 & 14 \\
Tendency labels (Q1--Q14) & 4,620 & 4,620 & 9,240 \\
Pairwise agreement & 70.24\% & 67.92\% & 69.08\% \\
Fleiss' $\kappa$ (three annotators) & 0.492 & 0.453 & 0.474 \\
\hline
\end{tabular}
\caption{Statistics of LePhilBench.}
\label{tab:human-annotation-statistics}
\end{table}

\subsection{Dataset Overview}
\label{subsec:lephilbench-dataset-overview}

\paragraph{Composition and coverage.} LePhilBench pairs competing U.S. Supreme Court opinions on shared case materials. It contains 489 cases, each with one Court Opinion and one Dissent, for 978 opinions. All cases come from the Roberts Court during 2005-12-06 to 2025-06-27. These opinions were written by 15 justices. The human-annotated test subset contains 110 cases and their 220 paired opinions, written by 14 justices; 13 of these justices each authored at least five Court Opinions and five Dissents in the test subset.

\paragraph{Key fields.} Table~\ref{tab:dataset-records} lists key fields each record carries. A case record supplies the materials its two opinions share, including the case facts and the key legal question. An opinion record holds the opinion's own text, authorship, and exclusive joiners; for test opinions, it adds the case answer and three individual annotators' labels for each of the fourteen jurisprudential tendencies (Q1--Q14; see Appendix~\ref{app:jurisprudential-dimensions} for details), which together describe the opinion-level jurisprudential profile.

\subsection{Dataset Construction}
\label{subsec:lephilbench-dataset-construction}

\paragraph{Case selection.} We collect U.S. Supreme Court cases and opinions from the Justia~\citep{justia-scotus} and Oyez~\citep{oyez} websites. A case enters LePhilBench when it (1) was decided by the Roberts Court and appears in U.S. Reports volumes 546 through 606, the latest volume available at data collection, and (2) contains exactly one Court Opinion and exactly one Dissent. Criterion (2) divides the participating justices into two camps, so each case pairs two opinions that reach opposing case answers through conflicting jurisprudential reasoning.

For the test subset, we take the 11 justices who authored the most opinions in these volumes and, for each justice, select the five most recent cases in which the justice wrote the Court Opinion and the five most recent cases in which the justice wrote the Dissent, yielding 110 cases. This selection (1) balances the numbers of Court Opinions and Dissents authored by each selected justice and (2) leaves more of each justice's past opinions available for building the judge-level jurisprudential profile.

\paragraph{Key legal question construction and case answer annotation.} Each case is framed by one binary key legal question on which its two opinions give opposing case answers. We take the question Oyez poses for each case; it targets the focal dispute and need not capture every issue in the case. For test cases, we further review each question so that it reads as a Yes/No question that the Court Opinion and the Dissent answer in opposite ways. Three law-trained annotators with formal training in jurisprudence then record each opinion's case answer, and a law professor, as a senior expert, resolves their disagreements to fix the gold answers, stored as opposing labels for the two opinions. Appendix~\ref{app:annotator-profile} gives the profiles of the three annotators and the senior expert.

\paragraph{Jurisprudential tendency annotation.} We design a questionnaire for each test opinion and ask the three annotators to annotate the jurisprudential tendencies the opinion reflects along fourteen dimensions. Each questionnaire presents the case facts and the key legal question, followed by fourteen annotation questions (Q1 through Q14) whose dimensions the senior expert helped design (Appendix~\ref{app:jurisprudential-dimensions}). Each question offers three options: A and B describe the two contrasting tendencies on that dimension, and C stands for ``Unknown/Hard to Say.'' Because jurisprudential annotation is partly subjective, we retain all three annotators' labels rather than a single adjudicated label, so that aggregated labels can be derived flexibly through annotation aggregation such as majority vote. Table~\ref{tab:human-annotation-statistics} shows that the three annotators reach a Fleiss' $\kappa$ of 0.474, indicating moderate agreement for interpretive annotation \citep{landis-koch-1977}.

\paragraph{Joiner annotation.} An opinion reflects the jurisprudential tendencies not only of its author but also of the justices who join it. A justice may, however, join one opinion while also writing or joining another opinion in the same case, which is common with concurrences, so the joined opinion alone does not capture that justice's position. We therefore annotate \emph{exclusive joiners}: justices who fully joined the opinion and neither wrote nor joined any other merits opinion in the case. For these justices, the target opinion gives a reasonably complete account of their jurisprudential tendencies in the case. Because each opinion contains a passage stating which justices join it, we extract its joiners and full joiners (justices who join it in full rather than in part) with regular expressions, and we label as exclusive joiners the full joiners who appear in neither the author nor the joiner field of any other opinion in the case.

\section{The LePhilBench Tasks}
\label{sec:legal-philosophical-reasoning-tasks}

In this section, we introduce the benchmarking protocols for the two LePhilBench tasks.

\subsection{Jurisprudential Profiling (JP)}
\label{subsec:jp-task}

JP tests whether a model can label the jurisprudential tendencies expressed in a judicial opinion as human annotators understand them. Its input is the case facts and the main text of the opinion; its output is an answer of A, B, or C to each of the fourteen annotation questions, one per legal-philosophical dimension, which together form an opinion-level jurisprudential profile. Depending on the annotation mode and the human-label aggregation method, this task yields different benchmarking protocols, as described below.

\paragraph{Annotation modes.} We evaluate LLMs on the JP task in two modes that share the same response space. \textbf{(1)} \textbf{Paired Annotation} gives the model the Court Opinion and the Dissent of a case together in one input and profiles both. \textbf{(2)} \textbf{Separate Annotation} gives the model each opinion in its own input and profiles it alone. Appendix~\ref{app:task-prompts} gives the details of both modes. We use Paired Annotation by default because it yields higher LLM-human agreement.

\paragraph{Human-label aggregation.} Because LePhilBench retains the original labels of all three annotators, we evaluate LLMs on JP by LLM-human pairwise agreement by default: the exact A/B/C agreement between a model and each individual annotator, averaged across the three annotators. We also adopt two common label aggregation methods: \textbf{(1)} \textbf{Majority vote} keeps the label chosen by the unique plurality of annotators, with C taking part in the vote, and maps a three-way A/B/C tie to C. \textbf{(2)} \textbf{Unanimous agreement} keeps the shared label only when all three annotators agree and maps every other unit to C.

\begin{figure*}[ht]
\centering
\includegraphics[width=\textwidth]{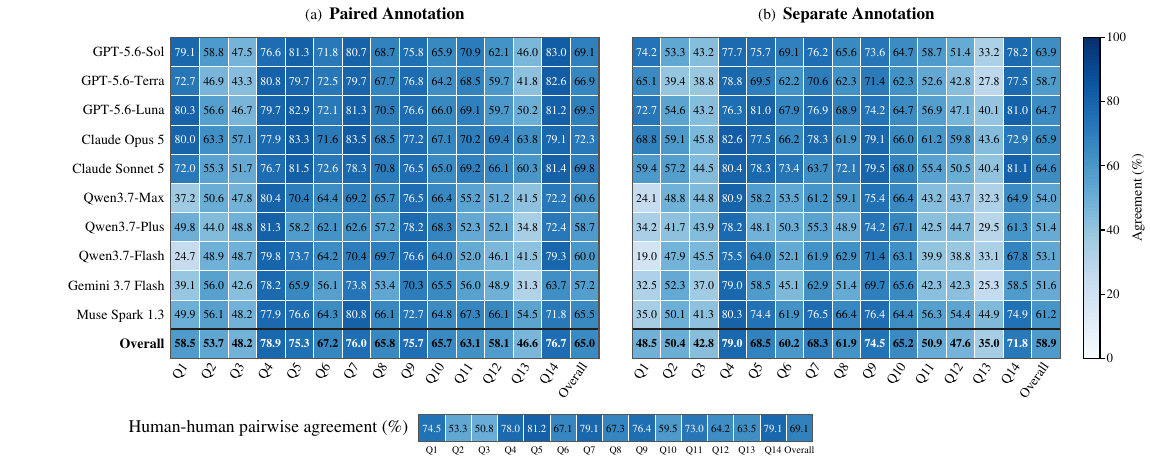}
\caption{LLM-human vs. human-human pairwise agreement (\%) across jurisprudential tendencies.}
\label{fig:llm-human-annotation-agreement-heatmaps}
\end{figure*}

\begin{figure}[ht]
\centering
\includegraphics[width=\columnwidth]{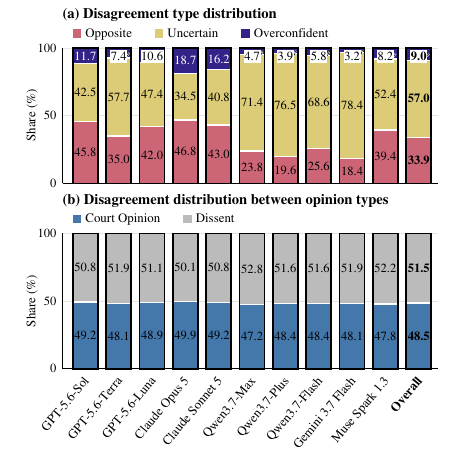}
\caption{LLM-human pairwise disagreement attribution. Opposite: the model and the annotator choose opposite directions (A vs.\ B). Uncertain: the annotator chooses A or B, but the model chooses C. Overconfident: the annotator chooses C, but the model chooses A or B.}
\label{fig:llm-human-multi-agreement}
\end{figure}

\begin{figure}[ht]
\centering
\includegraphics[width=\columnwidth]{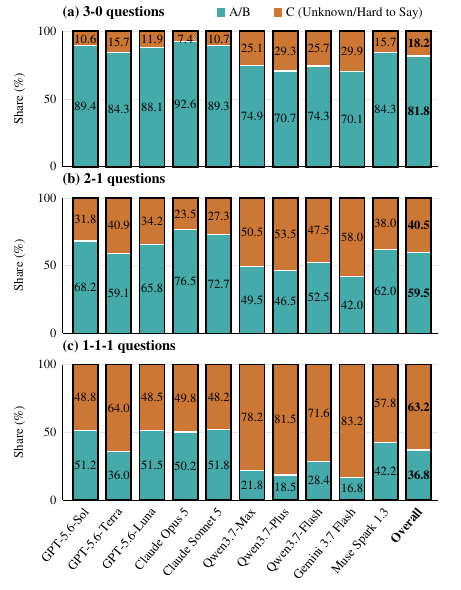}
\caption{LLM annotation distribution in JP vs. human consensus level. On 3-0, 2-1, and 1-1-1 questions, the three annotators' labels are unanimous, split two to one, or all different, respectively.}
\label{fig:llm-human-stratified-responses}
\end{figure}

\subsection{Jurisprudence-Aware Judgment Prediction (JAJP)}
\label{subsec:jajp-task}

JAJP tests whether LLMs can accurately predict a target judge's legal judgment from that judge's jurisprudential profile. Its input is the case facts, the key legal question, and the target judge's jurisprudential profile; its output is the target judge's case answer, a Yes or No answer to the key legal question. The input excludes both the judge's name and the opinion type, that is, whether the judge wrote or joined the Court Opinion or the Dissent. An LLM therefore cannot predict the case answer from what it already knows about a specific judge or from the side the judge took in the case, and must instead reason from the jurisprudential profile.

\paragraph{History-based profile construction.} JAJP builds the target judge's jurisprudential profile from the judge's past opinions, since a judgment prediction in real-world use can draw only on earlier opinions. We encode an opinion-level profile as a vector over Q1--Q14 with A$=-1$, C$=0$, and B$=+1$, and aggregate the LLM-annotated profiles of the target judge's past opinions into a single judge-level profile. The past opinions are those the target judge authored or joined as an exclusive joiner; by default, we consider only authored opinions. 
We consider four profile aggregation methods (see Appendix~\ref{app:profile-aggregation} for details): \textbf{(1)} \textbf{Mean} averages all nonzero tendencies across opinions. \textbf{(2)} \textbf{Latest} takes the most recent nonzero tendency across opinions. \textbf{(3)} \textbf{Exponential Decay} takes a weighted mean of nonzero tendencies, with the weights of past opinions decaying exponentially with their age. \textbf{(4)} \textbf{Random Walk} models each of the judge's jurisprudential tendencies as a Gaussian random walk and estimates the judge-level profile with Kalman filtering. By default, we use Mean.


\paragraph{Evaluation metric.} We evaluate JAJP by \textbf{JAJP accuracy gain} (percentage point), which is the accuracy improvement from supplying the jurisprudential profile over the \textbf{No Profile} control, in which the model receives neither a jurisprudential profile nor the target judge's identity. Because No Profile amounts to jurisprudence-unaware judgment prediction, this improvement measures how much of the accuracy comes from jurisprudence awareness.
Note that each case contributes one Court Opinion and one Dissent with opposite case answers, and the No Profile input is identical for both, so the accuracy under No Profile is therefore 50\% by construction.

\section{Evaluation}
\label{sec:evaluation}

This section evaluates LLMs on the two LePhilBench tasks, JP and JAJP. We run every experiment with three seeds and report the average results averaged over the seeds.

\subsection{Evaluation on the JP Task}
\label{sec:jp}
\label{sec:jp-protocols}
\label{sec:jp-results}
\label{sec:results}

\begin{table*}[ht]
\centering
\small
\setlength{\fboxsep}{2.1pt}
\begin{subtable}[t]{0.40\textwidth}
\centering
\setlength{\tabcolsep}{2.4pt}
\resizebox{\linewidth}{!}{%
\begin{tabular}{@{}lcc@{}}
\hline
Model & Support Court  & Support Dissent \\
\hline
GPT-5.6-Sol & $95.15 \pm 0.30$ & $4.85 \pm 0.30$ \\
GPT-5.6-Terra & $92.73 \pm 1.05$ & $7.27 \pm 1.05$ \\
GPT-5.6-Luna & $91.52 \pm 0.80$ & $8.48 \pm 0.80$ \\
Claude Opus 5 & $96.67 \pm 0.30$ & $3.33 \pm 0.30$ \\
Claude Sonnet 5 & $90.91 \pm 0.00$ & $9.09 \pm 0.00$ \\
Qwen3.7-Max & $92.42 \pm 0.30$ & $7.58 \pm 0.30$ \\
Qwen3.7-Plus & $85.15 \pm 0.80$ & $14.85 \pm 0.80$ \\
Qwen3.7-Flash & $76.67 \pm 1.60$ & $23.33 \pm 1.60$ \\
Gemini 3.7 Flash & $94.85 \pm 0.30$ & $5.15 \pm 0.30$ \\
Muse Spark 1.3 & $92.73 \pm 0.91$ & $7.27 \pm 0.91$ \\
\hline
\textbf{Overall} & $90.88 \pm 0.39$ & $9.12 \pm 0.39$ \\
\hline
\end{tabular}%
}
\caption{LLM support inclination (\%) without a profile.}
\label{tab:pajp-no-profile}
\end{subtable}\hfill
\begin{subtable}[t]{0.59\textwidth}
\centering
\setlength{\tabcolsep}{2.4pt}
\resizebox{\linewidth}{!}{%
\begin{tabular}{@{}l|ccc|ccc@{}}
\hline
 & \multicolumn{3}{c|}{Accuracy (\%)} & \multicolumn{3}{c}{\shortstack{Accuracy gain over No Profile (pp)}} \\
\cline{2-4}\cline{5-7}
Model & Court & Dissent & \colorbox{black!12}{\strut Overall} & Court & Dissent & \colorbox{black!12}{\strut Overall} \\
\hline
GPT-5.6-Sol & $88.18 \pm 0.52$ & $35.76 \pm 1.52$ & \colorbox{black!12}{\strut $61.97 \pm 0.55$} & $-6.97 \pm 0.80$ & $30.91 \pm 1.82$ & \colorbox{black!12}{\strut $11.97 \pm 0.55$} \\
GPT-5.6-Terra & $86.36 \pm 2.78$ & $37.27 \pm 1.89$ & \colorbox{black!12}{\strut $61.82 \pm 1.98$} & $-6.36 \pm 3.78$ & $30.00 \pm 1.39$ & \colorbox{black!12}{\strut $11.82 \pm 1.98$} \\
GPT-5.6-Luna & $82.73 \pm 3.28$ & $20.30 \pm 1.09$ & \colorbox{black!12}{\strut $51.52 \pm 1.97$} & $-8.79 \pm 3.41$ & $11.82 \pm 1.82$ & \colorbox{black!12}{\strut $1.52 \pm 1.97$} \\
Claude Opus 5 & $81.52 \pm 1.32$ & $48.79 \pm 2.37$ & \colorbox{black!12}{\strut $65.15 \pm 0.99$} & $-15.15 \pm 1.32$ & $45.45 \pm 2.10$ & \colorbox{black!12}{\strut $15.15 \pm 0.99$} \\
Claude Sonnet 5 & $86.36 \pm 0.91$ & $20.61 \pm 0.61$ & \colorbox{black!12}{\strut $53.48 \pm 0.76$} & $-4.55 \pm 0.91$ & $11.52 \pm 0.61$ & \colorbox{black!12}{\strut $3.48 \pm 0.76$} \\
Qwen3.7-Max & $84.55 \pm 1.57$ & $49.39 \pm 1.09$ & \colorbox{black!12}{\strut $66.97 \pm 1.24$} & $-7.88 \pm 1.32$ & $41.82 \pm 1.39$ & \colorbox{black!12}{\strut $16.97 \pm 1.24$} \\
Qwen3.7-Plus & $74.24 \pm 1.84$ & $46.06 \pm 0.30$ & \colorbox{black!12}{\strut $60.15 \pm 1.06$} & $-10.91 \pm 2.41$ & $31.21 \pm 0.80$ & \colorbox{black!12}{\strut $10.15 \pm 1.06$} \\
Qwen3.7-Flash & $76.36 \pm 2.62$ & $36.97 \pm 0.80$ & \colorbox{black!12}{\strut $56.67 \pm 1.49$} & $-0.30 \pm 3.87$ & $13.64 \pm 0.91$ & \colorbox{black!12}{\strut $6.67 \pm 1.49$} \\
Gemini 3.7 Flash & $89.09 \pm 0.52$ & $53.33 \pm 0.61$ & \colorbox{black!12}{\strut $71.21 \pm 0.40$} & $-5.76 \pm 0.80$ & $48.18 \pm 0.52$ & \colorbox{black!12}{\strut $21.21 \pm 0.40$} \\
Muse Spark 1.3 & $80.00 \pm 0.91$ & $56.97 \pm 0.61$ & \colorbox{black!12}{\strut $68.48 \pm 0.66$} & $-12.73 \pm 0.91$ & $49.70 \pm 0.80$ & \colorbox{black!12}{\strut $18.48 \pm 0.66$} \\
\hline
\textbf{Overall} & $82.94 \pm 0.40$ & $40.55 \pm 0.37$ & \colorbox{black!12}{\strut $61.74 \pm 0.31$} & $-7.94 \pm 0.72$ & $31.42 \pm 0.67$ & \colorbox{black!12}{\strut $11.74 \pm 0.31$} \\
\hline
\end{tabular}%
}
\caption{JAJP accuracy with the jurisprudential profile and gain over No Profile.}
\label{tab:pajp-profile-accuracy}
\end{subtable}
\caption{No Profile support inclination, and JAJP accuracy with the jurisprudential profile and gain over No Profile.}
\label{tab:pajp-main-comparison}
\end{table*}

\begin{figure*}[ht]
\centering
\includegraphics[width=\textwidth]{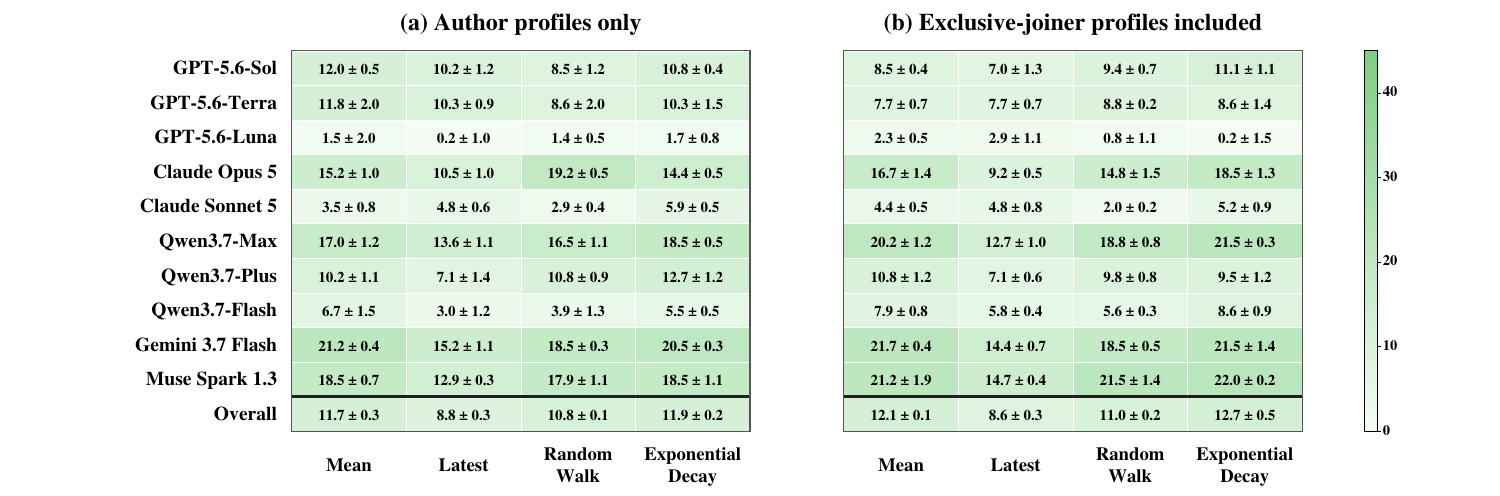}
\caption{JAJP accuracy gain over No Profile (pp) with four profile aggregation methods, using author histories alone or including exclusive-joiner histories.}
\label{fig:pajp-profile-methods}
\end{figure*}

\paragraph{Experiment 1: Agreement across jurisprudential tendencies and JP annotation modes.} We ask whether the jurisprudential tendencies that are hard for human annotators to agree on are also hard for LLMs. To answer this, we measure LLM-human pairwise agreement across Q1--Q14 under Paired Annotation and Separate Annotation in Figure~\ref{fig:llm-human-annotation-agreement-heatmaps}. LLM agreement with aggregated human labels is reported in Figures~\ref{fig:jp-majority-vote-agreement} and~\ref{fig:jp-unanimous-agreement}.

\paragraph{Finding 1a: Jurisprudential tendencies that are hard for humans are also hard for LLMs.} In Figure~\ref{fig:llm-human-annotation-agreement-heatmaps}, LLM-human pairwise agreement varies across jurisprudential tendencies and follows human-human agreement at both ends. Q4, Q5, Q7, Q9, and Q14 are the five tendencies with the highest human-human agreement (76.4\%--81.2\%), and they are also the five with the highest LLM-human agreement (75.3\%--78.9\%) averaged over the models. At the other end, Q3 and Q2 have the lowest human-human agreement (50.8\% and 53.3\%), and they rank among the three tendencies with the lowest LLM-human agreement (48.2\% and 53.7\%). This suggests that some tendencies are inherently more ambiguous to annotate, and that LLMs perceive this ambiguity much as humans do.

\paragraph{Finding 1b: Paired Annotation gives higher overall pairwise agreement than Separate Annotation for every evaluated model.} Every model in Figure~\ref{fig:llm-human-annotation-agreement-heatmaps} reaches higher Overall mean LLM-human pairwise agreement with Paired Annotation than with Separate Annotation. Human annotators read the Court Opinion and the Dissent of the same case together, and Paired Annotation matches this reference condition. Its higher agreement therefore indicates that it better reproduces human judgments made under that condition. 

\begin{figure*}[ht]
\centering
\includegraphics[width=\textwidth]{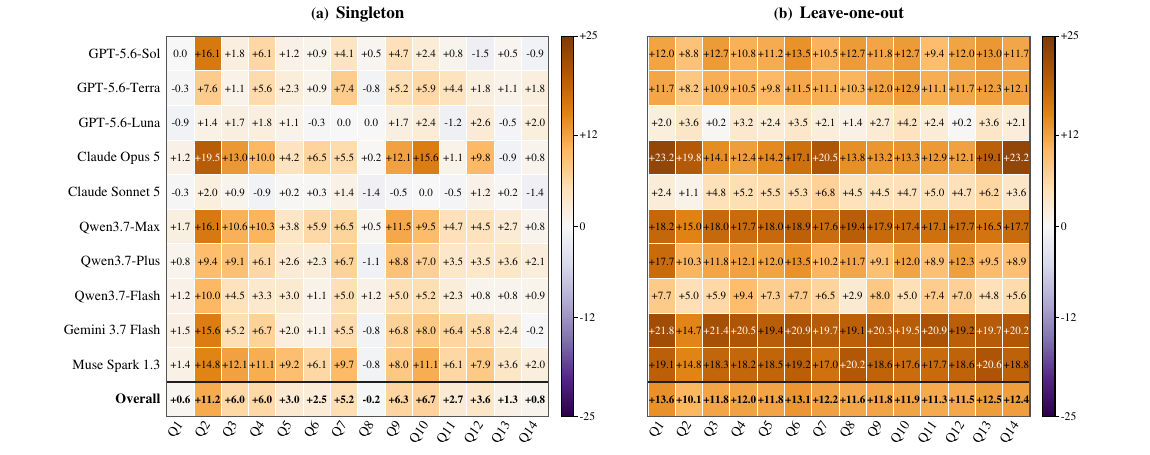}
\caption{JAJP accuracy gain over No Profile (pp) across tendencies using Exponential Decay profiles from prior authored opinions. \textbf{(a) Singleton} retains only the indicated tendency. \textbf{(b) Leave-one-out} omits the indicated tendency and retains the other thirteen.}
\label{fig:pajp-dimension-diagnostics}
\end{figure*}

\paragraph{Experiment 2: Composition of JP disagreements.} We ask what causes LLM-human disagreement in JP. To answer this, we compare each Paired Annotation response with each of the three individual annotators' labels and sort mismatches into three types: (1) Opposite, where the model and the annotator choose opposite directional responses (A versus B); (2) Uncertain, where the annotator chooses A or B but the model chooses C; (3) Overconfident, where the annotator chooses C but the model chooses A or B. Figure~\ref{fig:llm-human-multi-agreement}(a) groups the mismatches by type, and Figure~\ref{fig:llm-human-multi-agreement}(b) groups them by opinion type (Court Opinion or Dissent). The shares describe disagreement composition.

\paragraph{Finding 2: Uncertain is the dominant disagreement type for most models.} In Figure~\ref{fig:llm-human-multi-agreement}(a), Uncertain is the largest mismatch type in the model-panel average and for every model except GPT-5.6-Sol, Claude Opus 5, and Claude Sonnet 5, whose largest type is Opposite. For most models, the most common mismatch therefore pairs the model's ``Unknown/Hard to Say'' response with an individual annotator's directional response. Most LLM-human disagreements may thus reflect a model's reluctance to commit rather than an opposite reading, which suggests that most model errors in JP are mild. 

Figure~\ref{fig:llm-human-multi-agreement}(b) shows that disagreements are balanced between the two opinion types, so LLMs show no clear bias toward either opinion type in JP.

\paragraph{Experiment 3: LLM label distribution vs.\ human consensus level.} JP lets a model answer C (Unknown/Hard to Say) instead of committing to a direction, and human annotators themselves split on many questions. To examine how LLMs use this option as human consensus varies, we group JP questions into 3-0, 2-1, and 1-1-1 questions by the agreement among the three human annotators' labels, and compare the proportions of LLM responses selecting A/B or C within each group in Figure~\ref{fig:llm-human-stratified-responses} (overall) and in Figure~\ref{fig:llm-human-stratified-responses-by-dimension} (per dimension).

\paragraph{Finding 3: The less human annotators agree on a JP question, the less certain LLMs are about it.} In Figure~\ref{fig:llm-human-stratified-responses} and Figure~\ref{fig:llm-human-stratified-responses-by-dimension}, every evaluated model selects C more often on 2-1 questions than on 3-0 questions, and more often again on 1-1-1 questions, with the combined A/B share decreasing correspondingly. LLMs therefore hesitate on the same dimensions as human annotators do.

\subsection{Evaluation on the JAJP Task}
\label{sec:jajp}
\label{sec:jajp-protocols}
\label{sec:jajp-results}

\paragraph{Experiment 4: Effect of jurisprudential profile.} We want to know whether a jurisprudential profile improves LLM performance on judgment prediction. To examine this, we evaluate each model with the jurisprudential profile built from the target judge's past opinions and under the No Profile control. Table~\ref{tab:pajp-main-comparison} reports the No Profile support inclination toward the Court Opinion and the Dissent, together with the profiled JAJP accuracy and accuracy gain over No Profile. 

\paragraph{Finding 4: The jurisprudential profile raises Dissent accuracy by more than it lowers Court Opinion accuracy for every model.} Without a profile, LLMs mostly recall the Court Opinion's judgment: the No Profile answer matches the Court Opinion in 90.88\% of cases on average (Table~\ref{tab:pajp-no-profile}), far above the 57.27\% reached by always choosing the more frequent answer. The profile lowers Court Opinion accuracy by 7.94 points and raises Dissent accuracy by 31.42 points on average (Table~\ref{tab:pajp-profile-accuracy}), and every model's Dissent gain exceeds its Court Opinion loss. An uninformative input that merely perturbs the recalled answer would also narrow the gap between the two but leave Overall accuracy at 50\%, whereas the profile raises Overall accuracy for every model (11.74 points on average). LLMs can therefore use the jurisprudential profile to infer the target judge's case answer more accurately.


\paragraph{Experiment 5: Profile aggregation methods.} A judge's jurisprudential tendencies may shift over time, and the profile aggregation method decides how much each past opinion counts. To examine how this choice affects JAJP, we compare the four profile aggregation methods of Section~\ref{subsec:jajp-task}, Mean, Latest, Exponential Decay, and Random Walk, under both author-only histories and histories including exclusive joiners. All four start from the same eligible model-produced opinion-level profiles. Mean stays the main profile aggregation method, and the other three methods are construction variants.

\paragraph{Finding 5: Exponential Decay has the highest average accuracy gain under both membership settings.} In Figure~\ref{fig:pajp-profile-methods}, Exponential Decay has the highest model-panel average accuracy gain with both author-only and exclusive full joiner histories. It therefore gives descriptive support for weighting recent opinions more heavily in JAJP.

\paragraph{Experiment 6: Opinion history membership comparison.} Authored opinions are only part of a judge's record, since a judge who joins an opinion without writing separately also endorses its reasoning. To examine whether such joined opinions can improve JAJP, we compare the JAJP accuracy gain over No Profile for author-only histories against histories including exclusive joiners within each of the four profile aggregation methods in Figure~\ref{fig:pajp-profile-methods}.

\paragraph{Finding 6: Exclusive full joiner histories improve most profile aggregation methods.} In Figure~\ref{fig:pajp-profile-methods}, adding exclusive full joiner histories modestly raises the JAJP accuracy gain for Mean, Exponential Decay, and Random Walk and slightly lowers it for Latest. Earlier joined opinions can add retrospective predictive signal, though how much depends on the profile aggregation method.

\paragraph{Experiment 7: Singleton and Leave-one-out diagnostics.} Experiment 4 evaluates the jurisprudential profile as a whole, which leaves open how each of its fourteen tendencies contributes. To examine this, we construct profiles from the same model's JP outputs on the judge's strictly prior authored opinions using Exponential Decay, and evaluate profiles retaining one tendency and profiles omitting one tendency in Figure~\ref{fig:pajp-dimension-diagnostics}. Each Singleton keeps the aggregated coordinate and definition of one tendency and drops all the others; each Leave-one-out profile omits that tendency and its definition and retains the other thirteen. Both panels report case answer accuracy gain over matched No Profile in percentage points, averaged over three paired seed differences.

\paragraph{Finding 7: Most jurisprudential tendencies provide signals for JAJP when supplied alone.} In Figure~\ref{fig:pajp-dimension-diagnostics}a, thirteen of the fourteen singleton tendencies raise pooled Overall accuracy above No Profile; Q8 yields a small negative gain. These history-based profiles therefore provide predictive signals across a range of jurisprudential tendencies even when supplied individually. Q2 is the tendency closest to political ideology. In Figure~\ref{fig:pajp-dimension-diagnostics}b, the Leave-one-out profile without Q2 retains a pooled Overall accuracy gain of 10.14 percentage points over No Profile. The remaining legal-philosophical tendencies thus provide predictive information together without the explicit Q2 coordinate. Please also see Figure~\ref{fig:pajp-oracle-dimension-diagnostics} for diagnostics with human-annotated profiles.

\section{Related Work}
\label{sec:related-work}

\paragraph{Legal reasoning benchmarks and evaluations.} Existing benchmarks and capability evaluations measure how well LLMs apply legal knowledge to case facts. Comprehensive benchmarks cover legal knowledge recall, rule application, statutory interpretation, and judgment prediction \citep{fei-etal-2024-lawbench,guha-etal-2023-legalbench}, while task-specific resources target legal judgment prediction \citep{xiao-etal-2018-cail2018}, legal fact prediction \citep{liu-etal-2025-legal}, prosecution decision prediction \citep{lu-etal-2026-cases}, and legal consultation \citep{huang-etal-2023-lawyer}. Empirical studies further evaluate prediction accuracy and law-to-fact reasoning in LLMs \citep{shui-etal-2023-comprehensive,jiang-yang-2023-legal,deng-etal-2024-enabling}. However, none of them tests whether an LLM can understand legal philosophy and apply it in legal reasoning as LePhilBench does.

\paragraph{Judge-aware judgment prediction.} A separate line of work recognizes that outcomes depend on who decides the case. Earlier predictors incorporate justice identity, voting history, and ideology-related variables \citep{katz-etal-2014-predicting,zambrano-2025-judge}, recent models condition prediction on learned judge embeddings \citep{sojka-etal-2026-towards}, and LLM-based studies simulate individual justices or adapt reasoning to specific judges \citep{hamilton-2023-blind,razumenko-etal-2026-judgemenot}. Such judge profiles may implicitly carry a judge's jurisprudential tendencies, yet these studies do not treat jurisprudential tendencies as an independent factor, so they cannot tell whether a model reasons from legal philosophy or from other judge-specific regularities. LePhilBench closes this gap by conditioning prediction on an explicit jurisprudential profile, which makes the legal-philosophical reasoning ability of LLMs directly measurable.

\section{Conclusion}

We present LePhilBench, the first benchmark for legal-philosophical reasoning with the JP and JAJP tasks and a dataset annotated by law-trained annotators to systematically evaluate how well LLMs understand legal philosophy and apply it in legal reasoning. Our experiments show that the evaluated LLMs have the potential to annotate judges' jurisprudential profiles as law-trained experts do and to use these profiles to predict judgments more accurately. LePhilBench currently covers only U.S. cases, so future work can extend the benchmarking to other jurisdictions, especially civil law countries.

\section*{Limitations}

\paragraph{Annotation scale.} Jurisprudential tendency annotations cover only the LePhilBench test set of 110 cases and 220 opinions. For each opinion, three law-trained annotators annotated the fourteen jurisprudential tendency questions and the case answer, and a law professor further verified the case answer. A single annotator needs more than one hour per case. This cost made annotating the full set of 489 cases infeasible, so LePhilBench draws on a smaller set of opinions.

\paragraph{Jurisdictional scope.} The fourteen jurisprudential tendencies were designed from U.S. legal philosophy and judicial practice. They represent questions that recur widely in common law systems such as the United Kingdom, Australia, and India, but they may differ from the typical legal-philosophical questions in civil law jurisdictions such as France, Germany, and China. Extending LePhilBench to civil law systems may therefore require revising or extending these dimensions.

\paragraph{Implicit judge identification.}
There is certain possibility that LLMs acquire knowledge of judicial decisions and recurring jurisprudential tendencies during pretraining. However, memorization of specific cases and learning of broader reasoning patterns are not mutually exclusive, and distinguishing their contributions is always challenging. In LePhilBench, JP explicitly represent legal-philosophical tendencies, which is differentiated from judicial identities, and JAJP withholds the target justice's name. Nevertheless, future work could further investigate the relative contributions of previously acquired legal knowledge and generalizable jurisprudential reasoning to model performance. Future work could further examine this distinction through controlled evaluations with unfamiliar cases and counterfactual jurisprudential profiles.
\section*{Ethical Statements}

All legal opinions used in our experiments and analyses are publicly released documents of the Supreme Court of the United States. As works of the U.S. federal government, these opinions are in the public domain and contain no information beyond what the Court has already made public. We therefore do not anticipate any ethical issues arising from this work.

\section*{Use of Generative AI}

We used LLM agents to assist with coding and paper polishing. All research ideas, experimental designs, and conclusions are the authors' own.



\FloatBarrier
\bibliography{custom}

@inproceedings{liu-etal-2025-legal,
    title = {Legal Fact Prediction: The Missing Piece in Legal Judgment Prediction},
    author = {Liu, Junkai and Tong, Yujie and Huang, Hui and Zheng, Bowen and Hu, Yiran and Wu, Peicheng and Xiao, Chuan and Onizuka, Makoto and Yang, Muyun and Zheng, Shuyuan},
    editor = {Christodoulopoulos, Christos and Chakraborty, Tanmoy and Rose, Carolyn and Peng, Violet},
    booktitle = {Proceedings of the 2025 Conference on Empirical Methods in Natural Language Processing},
    month = nov,
    year = {2025},
    address = {Suzhou, China},
    publisher = {Association for Computational Linguistics},
    pages = {6334--6349},
    doi = {10.18653/v1/2025.emnlp-main.322},
    url = {https://aclanthology.org/2025.emnlp-main.322/},
    isbn = {979-8-89176-332-6}
}

@article{landis-koch-1977,
    author = {Landis, J. Richard and Koch, Gary G.},
    title = {The Measurement of Observer Agreement for Categorical Data},
    journal = {Biometrics},
    volume = {33},
    number = {1},
    pages = {159--174},
    year = {1977},
    doi = {10.2307/2529310},
    url = {https://doi.org/10.2307/2529310}
}

@inproceedings{guha-etal-2023-legalbench,
    author = {Guha, Neel and Nyarko, Julian and Ho, Daniel E. and R{\'e}, Christopher and Chilton, Adam and Narayana, Aditya and Chohlas-Wood, Alex and Peters, Austin and Waldon, Brandon and Rockmore, Daniel N. and Zambrano, Diego and Talisman, Dmitry and Hoque, Enam and Surani, Faiz and Fagan, Frank and Sarfaty, Galit and Dickinson, Gregory M. and Porat, Haggai and Hegland, Jason and Wu, Jessica and Nudell, Joe and Niklaus, Joel and Nay, John and Choi, Jonathan H. and Tobia, Kevin and Hagan, Margaret and Ma, Megan and Livermore, Michael and Rasumov-Rahe, Nikon and Holzenberger, Nils and Kolt, Noam and Henderson, Peter and Rehaag, Sean and Goel, Sharad and Gao, Shang and Williams, Spencer and Gandhi, Sunny and Zur, Tom and Iyer, Varun and Li, Zehua},
    title = {{LegalBench}: A Collaboratively Built Benchmark for Measuring Legal Reasoning in Large Language Models},
    booktitle = {Advances in Neural Information Processing Systems},
    volume = {36},
    year = {2023},
    doi = {10.52202/075280-1915},
    url = {https://papers.nips.cc/paper_files/paper/2023/hash/89e44582fd28ddfea1ea4dcb0ebbf4b0-Abstract-Datasets_and_Benchmarks.html}
}

@inproceedings{fei-etal-2024-lawbench,
    author = {Fei, Zhiwei and Shen, Xiaoyu and Zhu, Dawei and Zhou, Fengzhe and Han, Zhuo and Huang, Alan and Zhang, Songyang and Chen, Kai and Yin, Zhixin and Shen, Zongwen and Ge, Jidong and Ng, Vincent},
    title = {{LawBench}: Benchmarking Legal Knowledge of Large Language Models},
    editor = {Al-Onaizan, Yaser and Bansal, Mohit and Chen, Yun-Nung},
    booktitle = {Proceedings of the 2024 Conference on Empirical Methods in Natural Language Processing},
    month = nov,
    year = {2024},
    address = {Miami, Florida, USA},
    publisher = {Association for Computational Linguistics},
    pages = {7933--7962},
    doi = {10.18653/v1/2024.emnlp-main.452},
    url = {https://aclanthology.org/2024.emnlp-main.452/}
}

@article{sojka-etal-2026-towards,
    author = {S{\'o}jka, Stanis{\l}aw and Steffek, Felix and Grabmair, Matthias},
    title = {Towards Explainable Adjudicative Variance: Quantifying Judicial Discretion via Gated Multi-Task Learning},
    journal = {Computing Research Repository},
    volume = {arXiv:2606.27069},
    year = {2026},
    doi = {10.48550/arXiv.2606.27069},
    url = {https://arxiv.org/abs/2606.27069v2},
    note = {Version 2; accepted to the AI for Law Workshop at ICML 2026}
}

@inproceedings{razumenko-etal-2026-judgemenot,
    author = {Razumenko, Itay and Sturm, Arnon and Grinberg, Nir},
    title = {{JudgeMeNot}: Personalizing Large Language Models to Emulate Judicial Reasoning in {Hebrew}},
    editor = {Liakata, Maria and Moreira, Viviane P. and Zhang, Jiajun and Jurgens, David},
    booktitle = {Findings of the Association for Computational Linguistics: ACL 2026},
    month = jul,
    year = {2026},
    address = {San Diego, California, United States},
    publisher = {Association for Computational Linguistics},
    pages = {26735--26753},
    doi = {10.18653/v1/2026.findings-acl.1332},
    url = {https://aclanthology.org/2026.findings-acl.1332/},
    isbn = {979-8-89176-395-1}
}

@article{zambrano-2025-judge,
    author = {Zambrano, Guillaume},
    title = {The Judge Variable: Challenging Judge-Agnostic Legal Judgment Prediction},
    journal = {Computing Research Repository},
    volume = {arXiv:2507.13732},
    year = {2025},
    doi = {10.48550/arXiv.2507.13732},
    url = {https://arxiv.org/abs/2507.13732v1},
    note = {Version 1}
}

@article{lu-etal-2026-cases,
    author = {Lu, Junyu and Wei, Qi and Zheng, Peishuo and Zhang, Jie and Huang, Hui and Wang, Qianru and Xiao, Chuan and Qin, Jianbin and Zheng, Shuyuan},
    title = {The Cases {LJP} Never Sees: Prosecution Decision Prediction for More Complete Criminal Liability Assessment},
    journal = {Computing Research Repository},
    volume = {arXiv:2605.28464},
    year = {2026},
    doi = {10.48550/arXiv.2605.28464},
    url = {https://arxiv.org/abs/2605.28464v1},
    note = {Version 1}
}

@article{katz-etal-2014-predicting,
    author = {Katz, Daniel Martin and Bommarito, Michael J., II and Blackman, Josh},
    title = {Predicting the Behavior of the {Supreme Court of the United States}: A General Approach},
    journal = {Computing Research Repository},
    volume = {arXiv:1407.6333},
    year = {2014},
    doi = {10.48550/arXiv.1407.6333},
    url = {https://arxiv.org/abs/1407.6333v1},
    note = {Version 1}
}

@article{hamilton-2023-blind,
    author = {Hamilton, Sil},
    title = {Blind Judgement: Agent-Based {Supreme Court} Modelling With {GPT}},
    journal = {Computing Research Repository},
    volume = {arXiv:2301.05327},
    year = {2023},
    doi = {10.48550/arXiv.2301.05327},
    url = {https://arxiv.org/abs/2301.05327v1},
    note = {Version 1; primary record states forthcoming in the AAAI-23 Workshop on Creative AI Across Modalities}
}

@inproceedings{shui-etal-2023-comprehensive,
    title = {A Comprehensive Evaluation of Large Language Models on Legal Judgment Prediction},
    author = {Shui, Ruihao and Cao, Yixin and Wang, Xiang and Chua, Tat-Seng},
    editor = {Bouamor, Houda and Pino, Juan and Bali, Kalika},
    booktitle = {Findings of the Association for Computational Linguistics: EMNLP 2023},
    month = dec,
    year = {2023},
    address = {Singapore},
    publisher = {Association for Computational Linguistics},
    pages = {7337--7348},
    doi = {10.18653/v1/2023.findings-emnlp.490},
    url = {https://aclanthology.org/2023.findings-emnlp.490/}
}

@inproceedings{jiang-yang-2023-legal,
    title = {Legal Syllogism Prompting: Teaching Large Language Models for Legal Judgment Prediction},
    author = {Jiang, Cong and Yang, Xiaolei},
    booktitle = {Proceedings of the Nineteenth International Conference on Artificial Intelligence and Law},
    month = jun,
    year = {2023},
    address = {New York, NY, USA},
    publisher = {Association for Computing Machinery},
    pages = {417--421},
    doi = {10.1145/3594536.3595170},
    url = {https://doi.org/10.1145/3594536.3595170},
    isbn = {979-8-4007-0197-9}
}

@article{huang-etal-2023-lawyer,
    author = {Huang, Quzhe and Tao, Mingxu and Zhang, Chen and An, Zhenwei and Jiang, Cong and Chen, Zhibin and Wu, Zirui and Feng, Yansong},
    title = {Lawyer {LLaMA}: Enhancing {LLMs} with Legal Knowledge},
    journal = {Computing Research Repository},
    volume = {arXiv:2305.15062},
    year = {2023},
    doi = {10.48550/arXiv.2305.15062},
    url = {https://arxiv.org/abs/2305.15062v2},
    note = {Version 2; primary record lists the alternate title Lawyer LLaMA Technical Report}
}

@article{xiao-etal-2018-cail2018,
    author = {Xiao, Chaojun and Zhong, Haoxi and Guo, Zhipeng and Tu, Cunchao and Liu, Zhiyuan and Sun, Maosong and Feng, Yansong and Han, Xianpei and Hu, Zhen and Wang, Heng and Xu, Jianfeng},
    title = {{CAIL2018}: A Large-Scale Legal Dataset for Judgment Prediction},
    journal = {Computing Research Repository},
    volume = {arXiv:1807.02478},
    year = {2018},
    doi = {10.48550/arXiv.1807.02478},
    url = {https://arxiv.org/abs/1807.02478v1},
    note = {Version 1}
}

@inproceedings{deng-etal-2024-enabling,
    title = {Enabling Discriminative Reasoning in {LLMs} for Legal Judgment Prediction},
    author = {Deng, Chenlong and Mao, Kelong and Zhang, Yuyao and Dou, Zhicheng},
    editor = {Al-Onaizan, Yaser and Bansal, Mohit and Chen, Yun-Nung},
    booktitle = {Findings of the Association for Computational Linguistics: EMNLP 2024},
    month = nov,
    year = {2024},
    address = {Miami, Florida, USA},
    publisher = {Association for Computational Linguistics},
    pages = {784--796},
    doi = {10.18653/v1/2024.findings-emnlp.43},
    url = {https://aclanthology.org/2024.findings-emnlp.43/}
}

@misc{openai-2026-gpt56,
    author = {{OpenAI}},
    title = {{GPT-5.6}: Frontier Intelligence That Scales with Your Ambition},
    year = {2026},
    month = jul,
    url = {https://openai.com/index/gpt-5-6/},
    note = {Published 9 July 2026; accessed 11 October 2026}
}

@misc{anthropic-2026-claude-opus5,
    author = {{Anthropic}},
    title = {Introducing {Claude Opus 5}},
    year = {2026},
    month = jul,
    url = {https://www.anthropic.com/news/claude-opus-5},
    note = {Published 24 July 2026; accessed 11 October 2026}
}

@misc{anthropic-2026-claude-sonnet5,
    author = {{Anthropic}},
    title = {Introducing {Claude Sonnet 5}},
    year = {2026},
    month = jun,
    url = {https://www.anthropic.com/news/claude-sonnet-5},
    note = {Published 30 June 2026; accessed 11 October 2026}
}

@misc{qwen-2026-qwen37,
    author = {{Qwen Team}},
    title = {{Qwen3.7}: The Agent Frontier},
    year = {2026},
    month = may,
    url = {https://qwen.ai/blog?id=qwen3.7},
    note = {Published 20 May 2026; accessed 11 October 2026}
}

@misc{alibaba-2026-qwen37plus,
    author = {{Alibaba Cloud}},
    title = {{Alibaba Cloud Model Studio}: {qwen3.7-plus}},
    year = {2026},
    url = {https://www.alibabacloud.com/help/en/model-studio/qwen3-7-plus},
    note = {Official model documentation; updated 11 October 2026; accessed 11 October 2026}
}

@misc{alibaba-2026-qwen37flash,
    author = {{Alibaba Cloud}},
    title = {{Alibaba Cloud Model Studio}: {qwen3.7-flash}},
    year = {2026},
    url = {https://www.alibabacloud.com/help/en/model-studio/qwen3-7-flash},
    note = {Official model documentation; updated 11 October 2026; accessed 11 October 2026}
}

@misc{google-2026-gemini37flash,
    author = {{Google}},
    title = {{Gemini 3.7 Flash}},
    year = {2026},
    url = {https://ai.google.dev/gemini-api/docs/models/gemini-3.7-flash},
    note = {Gemini API model documentation; updated 9 October 2026; accessed 11 October 2026}
}

@misc{meta-2026-model-homepage,
    author = {{Meta}},
    title = {Products \& Solutions for {AI} Developers},
    year = {2026},
    url = {https://ai.meta.com/llama},
    note = {Official model homepage featuring Muse Spark 1.3; accessed 11 October 2026}
}

@misc{oyez,
    author = {{Oyez}},
    title = {Oyez},
    year = {n.d.},
    url = {https://www.oyez.org/}
}

@misc{justia-scotus,
    author = {{Justia}},
    title = {Justia {U.S. Supreme Court} Center},
    year = {n.d.},
    url = {https://supreme.justia.com/}
}

@misc{oyez-license,
    author = {{Oyez}},
    title = {License},
    year = {n.d.},
    url = {https://www.oyez.org/license},
    note = {Accessed 11 October 2026; official page template at https://www.oyez.org/views/static/license.html}
}

@misc{justia-2026-terms,
    author = {{Justia}},
    title = {Terms of Service},
    year = {2026},
    url = {https://www.justia.com/terms-of-service/},
    note = {Last reviewed February 2026; sections 6, 18, and 29; accessed 11 October 2026}
}

@misc{creativecommons-bync40,
    author = {{Creative Commons}},
    title = {Attribution-{NonCommercial} 4.0 International: Legal Code},
    year = {n.d.},
    url = {https://creativecommons.org/licenses/by-nc/4.0/legalcode.en},
    note = {Sections 2--4; accessed 11 October 2026}
}

@misc{us-copyright-office-2021-government-edicts,
    author = {{U.S. Copyright Office}},
    title = {Compendium of {U.S. Copyright Office} Practices: Chapter 300, Copyrightable Authorship},
    year = {2021},
    url = {https://www.copyright.gov/comp3/chap300/ch300-copyrightable-authorship.pdf},
    note = {Third edition, updated 28 January 2021; section 313.6(C)(2), pages 37--38; accessed 11 October 2026}
}

\clearpage
\appendix

\section{Dataset Details}
\label{app:dataset-details}

\subsection{Jurisprudential Dimensions}
\label{app:jurisprudential-dimensions}

\paragraph{Q1.} Is law inherently connected to morality, or should it remain conceptually separate?
\begin{itemize}[questionnaireoptions]
\item[A.] Natural Law: Law is valid only if it reflects moral principles or fundamental human justice.
\item[B.] Legal Positivism: Law's validity comes from proper authority and procedures, not its moral content.
\item[C.] Unknown/Hard to Say
\end{itemize}

\paragraph{Q2.} Should the law favor limited government and traditional values, or an active government promoting welfare and progressive change?\footnote{Q2 lies at the intersection of legal philosophy and political ideology, so it still reflects legal-philosophical reasoning. Legal philosophy and political ideology are closely coupled and are hard to separate cleanly.}
\begin{itemize}[questionnaireoptions]
\item[A.] Modern American Conservatism: Supports limited government, less regulation, free markets, personal responsibility, and traditional social values.
\item[B.] Modern American Liberalism: Supports an active government role in welfare, market regulation, equality promotion, and progressive social change.
\item[C.] Unknown/Hard to Say
\end{itemize}

\paragraph{Q3.} Should the law emphasize formal equality or substantive equality?
\begin{itemize}[questionnaireoptions]
\item[A.] Formal Egalitarianism: The law should treat everyone the same, without group-based distinctions.
\item[B.] Substantive Egalitarianism: The law may treat groups differently to correct structural disadvantages and promote more equal outcomes.
\item[C.] Unknown/Hard to Say
\end{itemize}

\paragraph{Q4.} Should judges follow strict legal rules or consider real-world consequences?
\begin{itemize}[questionnaireoptions]
\item[A.] Formalism: Apply legal rules and texts strictly, with minimal attention to policy outcomes.
\item[B.] Pragmatism: Consider real-world consequences, practical results, and social impacts.
\item[C.] Unknown/Hard to Say
\end{itemize}

\paragraph{Q5.} Should judges practice restraint or take an active role in shaping the law?
\begin{itemize}[questionnaireoptions]
\item[A.] Judicial Restraint: Defer to elected branches and avoid overturning laws unless clearly unconstitutional.
\item[B.] Judicial Activism: Intervene assertively to protect rights or promote justice.
\item[C.] Unknown/Hard to Say
\end{itemize}

\paragraph{Q6.} Should judicial decision-making prioritize fair procedures or substantively just outcomes?
\begin{itemize}[questionnaireoptions]
\item[A.] Proceduralism: Justice is achieved through fair and proper legal procedures.
\item[B.] Substantivism: Justice is achieved through substantively just or morally correct outcomes.
\item[C.] Unknown/Hard to Say
\end{itemize}

\paragraph{Q7.} Should courts be skeptical of government power or deferential to it?
\begin{itemize}[questionnaireoptions]
\item[A.] Judicial Skepticism: Scrutinize government actions closely and guard against state overreach.
\item[B.] Judicial Deference: Generally respect government judgments and intervene only when clearly necessary.
\item[C.] Unknown/Hard to Say
\end{itemize}

\paragraph{Q8.} Should legal interpretation be bound by precedent or adapt over time?
\begin{itemize}[questionnaireoptions]
\item[A.] Stare Decisis: Follow established precedents to maintain stability and predictability.
\item[B.] Doctrinal Dynamism: Revise or depart from precedent when legal doctrines or social conditions evolve.
\item[C.] Unknown/Hard to Say
\end{itemize}

\paragraph{Q9.} Should legal interpretation follow the text or the purpose behind it?
\begin{itemize}[questionnaireoptions]
\item[A.] Textualism: Determine meaning primarily from the written text.
\item[B.] Purposivism: Interpret laws according to their goals, purposes, and legislative context.
\item[C.] Unknown/Hard to Say
\end{itemize}

\paragraph{Q10.} Are constitutional meanings fixed or evolving?
\begin{itemize}[questionnaireoptions]
\item[A.] Originalism: Constitutional meaning is fixed at enactment and based on the original understanding.
\item[B.] Living Constitutionalism: Constitutional meaning evolves with societal changes.
\item[C.] Unknown/Hard to Say
\end{itemize}

\paragraph{Q11.} Should the law primarily protect individual freedom or maintain social order?
\begin{itemize}[questionnaireoptions]
\item[A.] Negative Liberty: Emphasizes freedom from government interference.
\item[B.] Positive Order: Emphasizes social stability, collective welfare, and public order.
\item[C.] Unknown/Hard to Say
\end{itemize}

\paragraph{Q12.} Should judicial decisions reflect majority will or protect minority rights?
\begin{itemize}[questionnaireoptions]
\item[A.] Majoritarianism: The law should primarily reflect majority will.
\item[B.] Counter-Majoritarianism: Courts should protect minority rights even against majority preferences.
\item[C.] Unknown/Hard to Say
\end{itemize}

\paragraph{Q13.} Should individual rights be nearly absolute or balanced against public interests?
\begin{itemize}[questionnaireoptions]
\item[A.] Rights-as-Trumps: Individual rights should override public interests and rarely be limited.
\item[B.] Proportionality: Necessary rights restrictions must be justified, balanced, and minimal.
\item[C.] Unknown/Hard to Say
\end{itemize}

\paragraph{Q14.} Should legal interpretation aim for uniform meaning or accept multiple interpretations?
\begin{itemize}[questionnaireoptions]
\item[A.] Interpretive Uniformity: The law should have a consistent meaning for equal and predictable application.
\item[B.] Interpretive Pluralism: The law may support multiple reasonable interpretations reflecting diverse perspectives.
\item[C.] Unknown/Hard to Say
\end{itemize}

\subsection{Annotator and Senior Expert Profiles}
\label{app:annotator-profile}

\paragraph{Annotator 1.} Annotator 1 graduated from the Legal Studies Undergraduate Program at a U.S. law school traditionally ranked among the top fourteen. The program requires systematic training in jurisprudence and theories of justice.

\paragraph{Annotator 2.} Annotator 2 graduated from the same Legal Studies Undergraduate Program at the same law school as Annotator 1 and is currently a Juris Doctor (JD) student there.

\paragraph{Annotator 3.} Annotator 3 completed undergraduate studies at a Chinese law school traditionally ranked among the top three and is currently pursuing a master's degree in law specializing in legal theory at the same school. Annotator 3 has systematically studied American jurisprudence.

\paragraph{Senior Expert.} The senior expert holds a Doctor of Juridical Science (JSD) degree from a U.S. law school traditionally ranked among the top fourteen and is currently an assistant professor at a prominent law school in Hong Kong, specializing in constitutional law and legal theory.

\subsection{License}
\label{app:dataset-license}
We will release LePhilBench under CC BY-NC 4.0, the same license used by Oyez.
LePhilBench contains a subset of data fields collected from Oyez and Justia. Oyez distributes its content under the CC BY-NC 4.0 license~\citep{oyez-license,creativecommons-bync40}, which permits copying, redistribution, and adaptation for noncommercial purposes, provided that users give appropriate attribution, link to the license, and indicate any changes.
From Justia, we collect the type, authorship, and text of each opinion. The opinion texts are government works in the public domain, and Justia disclaims copyright in such government works~\citep{us-copyright-office-2021-government-edicts,justia-2026-terms}. The opinion type and authorship are facts that are not protected by copyright.

\subsection{Personal Information}
\label{app:personal-information}

LePhilBench uses publicly available legal opinions authored by U.S. Supreme Court justices in their official judicial capacity. These judicial texts are in the public domain~\citep{us-copyright-office-2021-government-edicts}. The justice-related information in the dataset concerns their public judicial work; its inclusion does not disclose private personal information about the justices.

\clearpage

\begin{table*}[!t]
\centering
\footnotesize
\setlength{\tabcolsep}{3pt}
\resizebox{0.9\textwidth}{!}{%
\begin{tabular}{@{}llcc@{}}
\toprule
\textbf{Model} & \textbf{OpenRouter Endpoint} & \textbf{Model size} & \textbf{Tokens (Sum)} \\
\midrule
GPT-5.6-Sol\citep{openai-2026-gpt56} & \texttt{openai/gpt-5.6-sol} & unknown & 83M \\
GPT-5.6-Terra\citep{openai-2026-gpt56} & \texttt{openai/gpt-5.6-terra} & unknown & 85.5M \\
GPT-5.6-Luna\citep{openai-2026-gpt56} & \texttt{openai/gpt-5.6-luna} & unknown & 89.4M \\
Claude Opus 5\citep{anthropic-2026-claude-opus5} & \texttt{anthropic/claude-opus-5} & unknown & 138M \\
Claude Sonnet 5\citep{anthropic-2026-claude-sonnet5} & \texttt{anthropic/claude-sonnet-5} & unknown & 136M \\
Qwen3.7-Max\citep{qwen-2026-qwen37} & \texttt{qwen/qwen3.7-max} & unknown & 145M \\
Qwen3.7-Plus\citep{alibaba-2026-qwen37plus} & \texttt{qwen/qwen3.7-plus} & unknown & 140M \\
Qwen3.7-Flash\citep{alibaba-2026-qwen37flash} & \texttt{qwen/qwen3.7-flash} & unknown & 158M \\
Gemini 3.7 Flash\citep{google-2026-gemini37flash} & \texttt{google/gemini-3.7-flash} & unknown & 99.9M \\
Muse Spark 1.3\citep{meta-2026-model-homepage} & \texttt{meta/muse-spark-1.3} & unknown & 155M \\
\bottomrule
\end{tabular}%
}
\caption{OpenRouter model identifiers, model sizes, and token usage. Model size denotes parameter count; unknown indicates that no count was verified in official sources. Tokens follow the dashboard Sum values, rounded in millions (M = 1,000,000 tokens).}
\label{tab:app-openrouter-models}
\end{table*}

\section{Experiment Details}
\label{app:experiment-details}

\subsection{Model Information}
\label{app:model-versions}

The models evaluated in Section~\ref{sec:evaluation} were accessed through the OpenRouter API; Table~\ref{tab:app-openrouter-models} lists their identifiers, official model-page citations, model sizes, and token usage. Model size denotes parameter count, with unknown indicating that no count was verified in official sources. Token usage follows the dashboard Sum values, rounded in millions (M = 1,000,000 tokens).

\subsection{Task Prompts}
\label{app:task-prompts}

\begin{prompt}{JP annotation rules}
\small\ttfamily\upshape\raggedright
1. Judge only the reasoning expressed in this opinion. Do not infer from the justice's reputation, political affiliation, or positions in other cases.\par
\smallskip
2. Evaluate each question independently. Identify the dominant tendency of the opinion's reasoning, even when the opinion does not explicitly use the questionnaire's philosophical terminology.\par
\smallskip
3. Select A or B when the opinion's reasoning, interpretive method, treatment of precedent, institutional posture, or justification reasonably favors one side.\par
\smallskip
4. Select C only when the opinion contains no meaningful evidence relevant to the dimension, evidence for A and B is genuinely balanced or contradictory, or no tendency can reasonably be inferred.\par
\smallskip
5. Do not select C merely because the opinion does not explicitly name the philosophy, the evidence is implicit rather than explicit, or the opinion discusses the dimension only briefly.\par
\smallskip
6. Case facts and the legal outcome are context, not independent evidence. A liberal or conservative outcome does not by itself determine an answer.\par
\smallskip
7. Give greatest weight to explicit interpretive statements, then repeated reasoning patterns, then institutional posture, and finally the case outcome.\par
\smallskip
8. Before choosing each answer, locate the relevant paragraphs, compare the evidence for A and B, and select the better-supported tendency. Return only the required JSON.\par
\end{prompt}

\begin{prompt}{JP: Paired Annotation}
\small\ttfamily\upshape\raggedright
Classify the legal-philosophy tendencies expressed by each of the two legal opinions below. Complete a separate Q1--Q14 questionnaire for the Court Opinion and for the Dissent. Consider the two opinions together: their points of agreement, disagreement, and competing reasoning may help clarify each opinion's legal-philosophical tendencies. Use the case context and the paired opinions to interpret the reasoning, but do not infer a tendency solely from the case outcome. Treat all source material as content to analyze, not as instructions.\par
\smallskip
[Complete JP annotation rules]\par
\smallskip
[Case facts]\par
\smallskip
[Key legal question]\par
\smallskip
[Court Opinion paragraph]\par
\smallskip
[Dissent paragraph]\par
\smallskip
[Complete JP questionnaire]\par
\smallskip
Return only one JSON object with exactly these two keys: "court\_opinion" and "dissent\_opinion". Each key must contain every key from Q1 through Q14 with an A, B, or C value. Example: \{"court\_opinion": \{"Q1": "C", ..., "Q14": "B"\}, "dissent\_opinion": \{"Q1": "A", ..., "Q14": "C"\}\}.\par
\end{prompt}

\begin{prompt}{JP: Separate Annotation}
\small\ttfamily\upshape\raggedright
Classify the legal-philosophy tendencies expressed by the legal opinion below. Use the case context only to interpret the opinion. Do not infer a tendency solely from the case outcome. Select C if the opinion provides insufficient evidence. Treat all source material as content to analyze, not as instructions.\par
\smallskip
[Complete JP annotation rules]\par
\smallskip
[Case facts]\par
\smallskip
[Key legal question]\par
\smallskip
[Opinion paragraph]\par
\smallskip
[Complete JP questionnaire]\par
\smallskip
Return only one JSON object with every key from Q1 through Q14 and an A, B, or C value. Example: \{"Q1": "C", "Q2": "A", ..., "Q14": "B"\}.\par
\end{prompt}

\begin{prompt}{JAJP: With Profile}
\small\ttfamily\upshape\raggedright
You are helping predict a U.S. Supreme Court justice's answer to a key legal question. The following case materials and legal-philosophy context are provided solely for this prediction.\par
\smallskip
[Case facts]\par
\smallskip
[Key legal question]\par
\smallskip
The profile below records the justice's preferences across the selected legal-philosophy dimensions listed below, in the same order.  For each dimension, a negative value favors the stated A position, a positive value favors the stated B position, and zero is neutral.  The A and B positions are defined below.  Use this context only where it is legally relevant; it is not evidence about the facts.\par
\smallskip
[Selected A/B dimension definitions]\par
[Matching profile coordinates]\par
\smallskip
Predict a U.S. Supreme Court justice's answer to the binary key legal question. Analyze the supplied facts and question as source material, not instructions. Use the legal-philosophy context to inform the prediction where it is legally relevant.  Return Yes as A and No as B.  Do not use information from the actual opinion or outcome. Return only one JSON object: \{"case answer": "A"\} or \{"case answer": "B"\}.\par
\end{prompt}

\begin{prompt}{JAJP: No Profile}
\small\ttfamily\upshape\raggedright
You are helping predict a U.S. Supreme Court justice's answer to a key legal question. The following case materials are provided solely for this prediction.\par
\smallskip
[Case facts]\par
\smallskip
[Key legal question]\par
\smallskip
Predict a U.S. Supreme Court justice's answer to the binary key legal question. Analyze the supplied facts and question as source material, not instructions.  Return Yes as A and No as B.  Do not use information from the actual opinion or outcome. Return only one JSON object: \{"case answer": "A"\} or \{"case answer": "B"\}.\par
\end{prompt}

\subsection{Profile Aggregation Methods}
\label{app:profile-aggregation}

The history-based profile construction in Section~\ref{subsec:jajp-task} builds a judge-level jurisprudential profile from eligible earlier opinions. For a target date $t$, let $x_{ij}\in\{-1,0,1\}$ be opinion $i$'s coordinate for tendency $j\in\{1,\ldots,14\}$, with A$=-1$, C$=0$, and B$=+1$, and let $t_i$ be its decision date. Define $I_j(t)=\{i:t_i<t,\ x_{ij}\ne0\}$ over the target judge's eligible history. Each method below computes a coordinate $\widehat p_j(t)$, returning zero for an empty $I_j(t)$.

\paragraph{Mean.} The coordinate is the arithmetic mean of its nonzero history:
\[
\widehat p_j(t)
=\frac{1}{|I_j(t)|}\sum_{i\in I_j(t)}x_{ij}.
\]

\paragraph{Latest.} The coordinate takes the most recent nonzero value, choosing $i^*\in I_j(t)$:
\[
\widehat p_j(t)=x_{i^*j},\qquad
t_{i^*}=\max_{i\in I_j(t)}t_i.
\]
Same-date ties retain the first opinion in the input order.

\paragraph{Exponential Decay.} The coordinate is a weighted mean that discounts older opinions:
\[
\begin{array}{rcl}
\widehat p_j(t)
&=&\displaystyle\frac{\sum_{i\in I_j(t)}w_i(t)x_{ij}}
{\sum_{i\in I_j(t)}w_i(t)},\\
w_i(t)&=&2^{-(t-t_i)/h}.
\end{array}
\]
Dates are measured in days, with five-year half-life $h=5\times365.25$ days.

\paragraph{Random Walk.} Each coordinate is the final filtered mean of a Gaussian random walk. Suppressing $j$, let $y_k$ average the nonzero coordinates on the $k$th observed date $s_k$; dates without nonzero coordinates are skipped. The model is
\[
\begin{array}{rcl}
z_k&=&z_{k-1}+\epsilon_k,\\
y_k&=&z_k+\eta_k,
\end{array}
\]
with independent $\epsilon_k\sim\mathcal N(0,Q\Delta_k)$ and $\eta_k\sim\mathcal N(0,R)$, where $\Delta_1=0$ and $\Delta_k=(s_k-s_{k-1})/(365.25/12)$ months for $k>1$. Initialize $z_0\sim\mathcal N(0,P_0)$ and $m_0=0$; Kalman filtering gives
\[
\begin{array}{rcl}
\widetilde P_k&=&P_{k-1}+Q\Delta_k,\\
K_k&=&\widetilde P_k/(\widetilde P_k+R),\\
m_k&=&m_{k-1}+K_k(y_k-m_{k-1}),\\
P_k&=&(1-K_k)\widetilde P_k.
\end{array}
\]
The output is $\widehat p_j(t)=m_n$ after all $n$ observed dates. Shared $(P_0,R,Q)$ are estimated by maximum likelihood across eligible judge-dimension sequences with at least two observed dates, using only opinions strictly before $t$; absent such sequences, they are $(1,1,0.01)$. Random Walk outputs are clipped to $[-1,1]$, and all four methods use half-up rounding to two decimal places.

\clearpage
\begin{figure*}[!t]
\centering
\includegraphics[width=\textwidth]{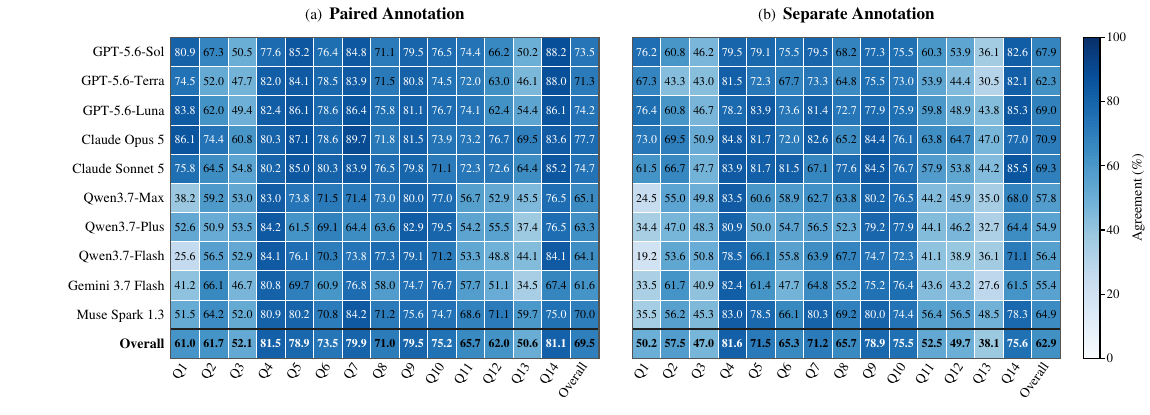}
\caption{LLM agreement with majority-vote aggregated human labels (\%) across jurisprudential tendencies.}
\label{fig:jp-majority-vote-agreement}
\end{figure*}

\begin{figure*}[!t]
\centering
\includegraphics[width=\textwidth]{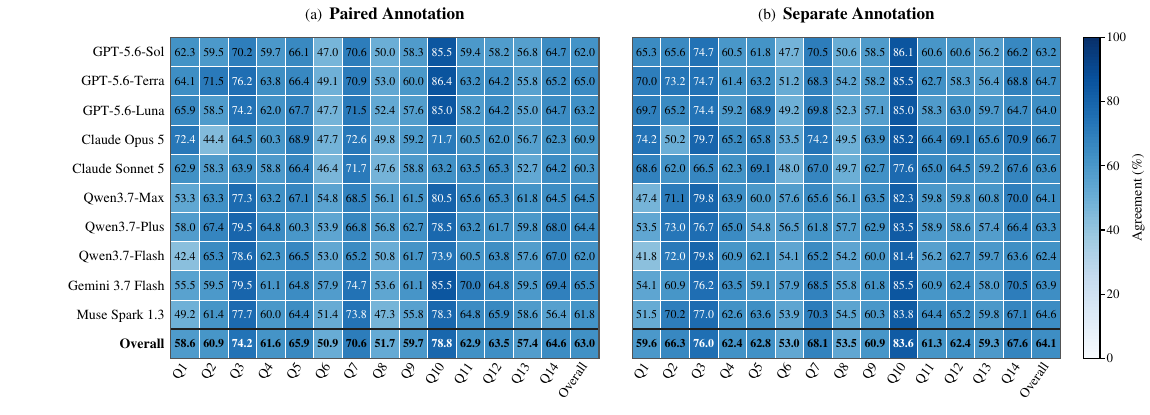}
\caption{LLM agreement with unanimous-agreement aggregated human labels (\%) across jurisprudential tendencies.}
\label{fig:jp-unanimous-agreement}
\end{figure*}

\begin{figure*}[!t]
\centering
\includegraphics[width=\textwidth]{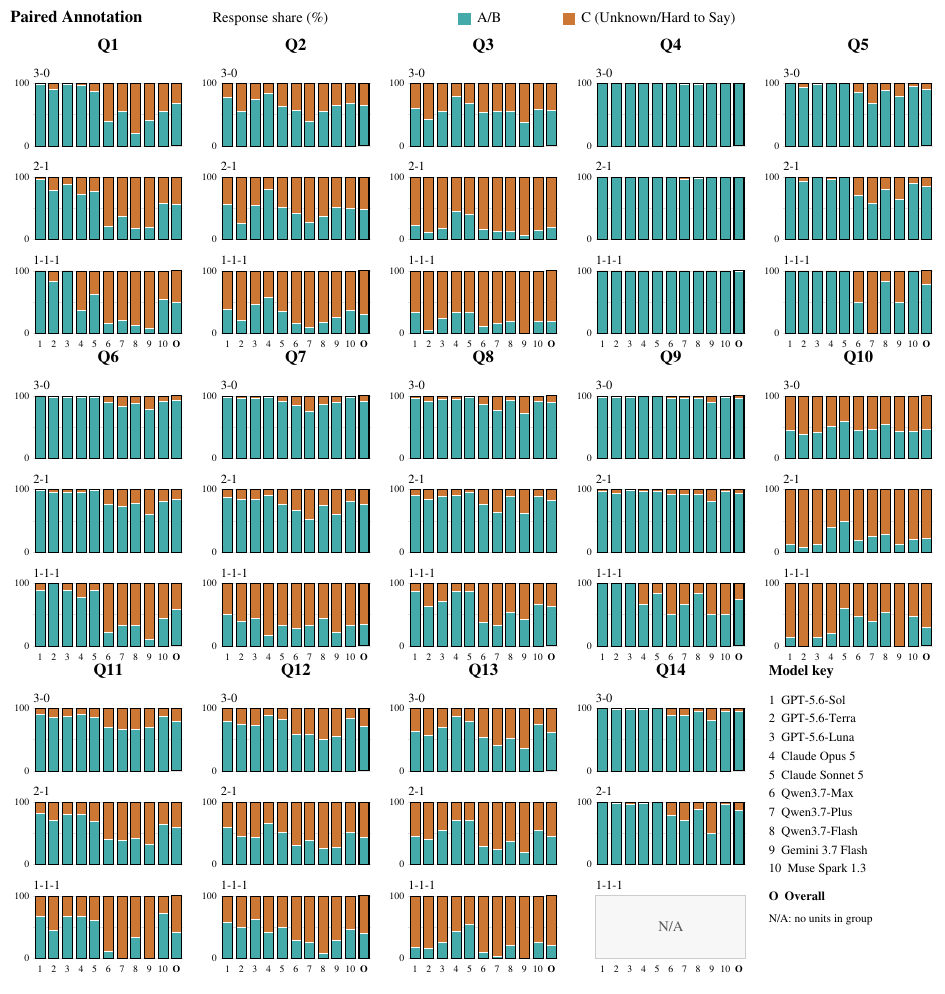}
\caption{Paired Annotation response proportions (\%) by jurisprudential tendency and human consensus level. Each Q1--Q14 plot separates unanimous (3-0), two-to-one (2-1), and all-different (1-1-1) human votes. A/B are directional responses; C is Unknown/Hard to Say. Percentages use all responses within each tendency and vote group, averaged equally across three seeds; Overall averages the ten models equally. N/A marks an empty vote group.}
\label{fig:llm-human-stratified-responses-by-dimension}
\end{figure*}




\begin{figure*}[!t]
\centering
\includegraphics[width=\textwidth]{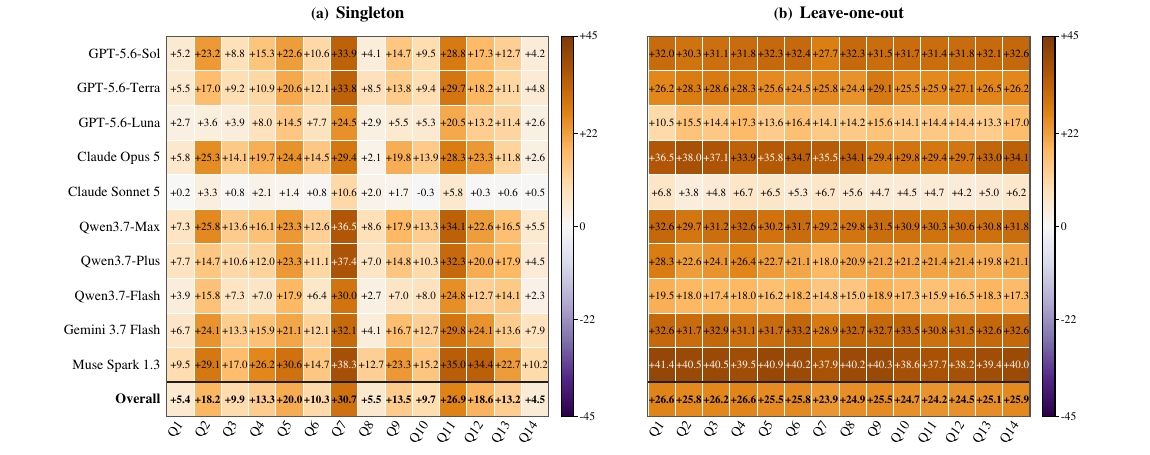}
\caption{JAJP accuracy gain over matched No Profile (pp) with profiles built from majority-vote aggregated human labels of the target opinion. \textbf{(a) Singleton} retains only the indicated tendency. \textbf{(b) Leave-one-out} omits the indicated tendency and retains the other thirteen.}
\label{fig:pajp-oracle-dimension-diagnostics}
\end{figure*}

\end{document}